\documentclass[smallcondensed]{svjour3}                     
\smartqed  
\usepackage{graphicx}

\usepackage{mathptmx}      

\usepackage{amssymb}
\usepackage{amsmath}
\usepackage{stmaryrd}
\usepackage{booktabs}
\usepackage{natbib}
\usepackage{color}
\usepackage{url}

\newcommand{\bx}{\vec{x}}
\newcommand{\by}{\vec{y}}
\newcommand{\bt}{\vec{t}}

\journalname{Data Mining and Knowledge Discovery}

\begin{document}

\title{Multi-target prediction: \\
A unifying view on problems and methods
}



\author{Willem Waegeman \and Krzysztof Dembczy\'nski \and Eyke H{\"u}llermeier}


\institute{
           W. Waegeman \at
              Department of Mathematical Modelling, Statistics and Bioinformatics, Ghent University,
              Coupure links 653, B-9000 Ghent,
              Belgium\\
				  \email{willem.waegeman@ugent.be}
					\and
					K. Dembczy\'nski \at
               Institute of Computing Science, Pozna\'n University of Technology, Piotrowo 2, 60-965~Pozna\'n,~Poland\\        
              \email{kdembczynski@cs.put.poznan.pl}   
           \and
           E. H{\"u}llermeier \at
              Department of Computer Science, Paderborn University,
              Pohlweg 49--51
33098 Paderborn, Germany \\
              \email{eyke@upb.de}           
					 }

\date{Received: date / Accepted: date}

\maketitle

\begin{abstract}
Multi-target prediction (MTP) is concerned with the simultaneous prediction of multiple target variables of diverse type. Due to its enormous application potential, it has developed into an active and rapidly expanding research field that combines several subfields of machine learning, including multivariate regression, multi-label classification, multi-task learning, dyadic prediction, zero-shot learning, network inference, and matrix completion. In this paper, we present a unifying view on MTP problems and methods. First, we formally discuss commonalities and differences between existing MTP problems. To this end, we introduce a general framework that covers the above subfields as special cases. As a second contribution, we provide a structured overview of MTP methods. This is accomplished by identifying a number of key properties, which distinguish such methods and determine their suitability for different types of problems. Finally, we also discuss a few challenges for future research.   


\keywords{multivariate regression, multi-label classification, multi-task learning, pairwise learning, dyadic prediction, zero-shot learning, collaborative filtering}
\end{abstract}

\section{Introduction}
\label{sec:introduction}

In contrast to conventional supervised learning, where a single target variable needs to be predicted on the basis of a set of features describing an instance, multi-target prediction (MTP) is concerned with the simultaneous prediction of multiple target variables of possibly different type, such as binary, nominal, ordinal, or real-valued. Applications of multi-target prediction are omnipresent in the digitalized society of the 21st century. Classical applications that are often studied in machine learning papers include image tagging in computer vision, document cateogorization in text mining, and product recommendation in online advertising. Besides, MTP problems arise in many other application domains as well.  In medicine, one is interested in predicting several clinical outcomes of patients at the same time. In climate sciences, one would like to forecast extreme weather events for many regions in the world, which are related through complex physical and geological processes. In biology, one would like to unravel the different biological functions that a gene might express. In chemistry, one would like to know which molecules might be potential drugs to cure a given disease. In social networks, one intends to predict which users interact with a given user. In ecology, one constructs species distribution models, which describe the prevalence of different types of species in a given habitat. 

Applications of that kind have resulted in novel research questions, and a need for developing specialized multi-target prediction methods. Often, several target variables are related to each other, for example because they obey certain constraints (for instance, if variables are positions in a ranking, they must be mutually exclusive), they correspond to nodes of a graph, or they provide evidence of parent-child relationships. In other situations, specific properties of the targets are known, such as molecular structures or feature representations. Obviously, knowledge of that kind could be used to improve predictive performance. However, in many applications, neither target relations nor representations are known a priori. Instead, they need to be discovered from the data. 
The main credo of research in MTP is the conviction that, compared to the most obvious approach of learning an individual prediction model for each target variable independently of the others, the exploitation of target dependencies will lead to better performance.

In this review-style article, we present a unifying view of MTP problems and methods. 
When speaking about multi-target prediction, machine learning researchers often allude to either multi-label classification, multi-output regression (a.k.a.\ multivariate regression in statistics), or multi-task learning problems.  Moreover, if additional side information in the form of target relations or target representations is accessible, those three settings further extend to multi-target prediction scenarios that are known as dyadic prediction, hierarchical multi-label classification, and zero-shot learning in the literature. Besides, there is a close connection to matrix completion and network inference, which becomes obvious when representing the relationship between instances and targets in the form of a matrix or a graph.   
Despite strong commonalities, there is little interaction between the different sub-communities. Moreover, there are several problems that have been studied in different communities under different names. Sometimes there is even terminological confusion within the same community. 

As a first contribution, we provide a structured overview of the multitude of MTP problems. To this end, we present a formal framework for multi-target prediction in Section~2. By identifying a set of characteristic properties, we subdivide MTP problems into a number of well-known settings. 
Particular attention will be paid to the formal definitions of problems that characterize the fields of multi-label classification, multivariate regression, multi-task learning, zero-shot learning, and matrix completion.  We also discuss a number of related settings, such as structured output prediction and multi-class classification, and argue why those settings should not be covered by the umbrella of MTP.  

Subsequently, in Section~3, we present a unifying view of MTP methods. 
Here, we intend to unravel a number of general mechanisms that are essential for obtaining state-of-the-art predictive performance in MTP. As will be seen, the applicability or usefulness of a method strongly depends on properties of the problem setting, such as whether or not side information is available for targets, and how this side information looks like. Another important question is whether or not one intends to generalize to novel targets and/or instances. 

The overall goal of this paper is to provide insight into the vast literature on MTP problems and methods, especially for readers who are new to the field.  The paper is not a typical review paper, however, and does not lay claim on being comprehensive in this regard---in light of the breadth of the field, that appears to be an impossible endeavor. 
Instead, we intend to focus on some general principles that might be helpful in identifying the right approach for a given problem. In Section~4, we conclude with a couple of remarks and challenges for future research, and briefly discuss some important aspects of multi-target prediction that are less emphasized in this paper.

\section{A unifying view on MTP problems}

\subsection{A general framework for multi-target prediction}

In this section, we establish links between different MTP problems. We start by describing a general framework that covers both simple and more advanced MTP problems. Formally, our framework is defined as follows.

\begin{definition} {\bf (Multi-target prediction)}
A multi-target prediction setting is characterized by instances $\vec{x} \in \mathcal{X}$ and targets $\vec{t} \in \mathcal{T}$ with the following properties: 
\begin{itemize} 
\item[P1.] A training dataset $\mathcal{D}$ consists of triplets $(\vec{x}_i,\vec{t}_j,y_{ij})$, where $y_{ij} \in \mathcal{Y}$ denotes a score that characterizes the relationship between the instance $\vec{x}_i$ and the target $\vec{t}_j$.  
\item[P2.] In total, $n$ different instances and $m$ different targets are observed during training, with $n$ and $m$ finite numbers. Thus, the scores $y_{ij}$ of the training data can be arranged in an $n \times m$ matrix $Y$, which is in general incomplete, i.e., $Y$ has missing values.
\item[P3.] The score set $\mathcal{Y}$ is one-dimensional. It consists of nominal, ordinal or real values.  
\item[P4.] The goal consists of predicting scores for any instance-target couple $(\vec{x},\vec{t}) \in \mathcal{X} \times \mathcal{T}$.   
\end{itemize}
\end{definition}

The above definition accommodates the availability of side information for targets. To keep notation simple, we stick to vector representations in our definitions, i.e., we identify targets with feature vectors $\vec{t}$. We remark, however, that other types of side knowledge, such as structured representations or relations, could also be considered. The examples to be discussed below will make this point more clear. 

In what follows, we show that various multi-target prediction problems are recovered as specific instantiations of the above framework. 
As already said, if side information for targets is available, we assume it can be encoded in the vector representation $\vec{t}$. If no side information is available, $\vec{t}$ will be an uninformative vector (e.g., merely consisting of an identifying number).

\subsection{Conventional multi-target prediction settings}

We start by explaining what we call conventional multi-target prediction problems. For those problems, side information for targets is normally not available. 
Multi-label classification, multivariate regression, and multi-task learning are the most well-known subfields of machine learning that can be mentioned as examples. Multivariate regression and multi-label classification consider the simultaneous prediction of multiple binary or real-valued targets, respectively~\citep{Demb2012a,Tsoumakas_and_Katalos_2007}. Multi-task learning then unifies those subfields, and further extends them to problems where not all targets are relevant for all instances~\citep{Caruana1997}. To make this point clearer, we discuss three prototypical examples.

\begin{figure}
\begin{center}
 \includegraphics[width=6cm]{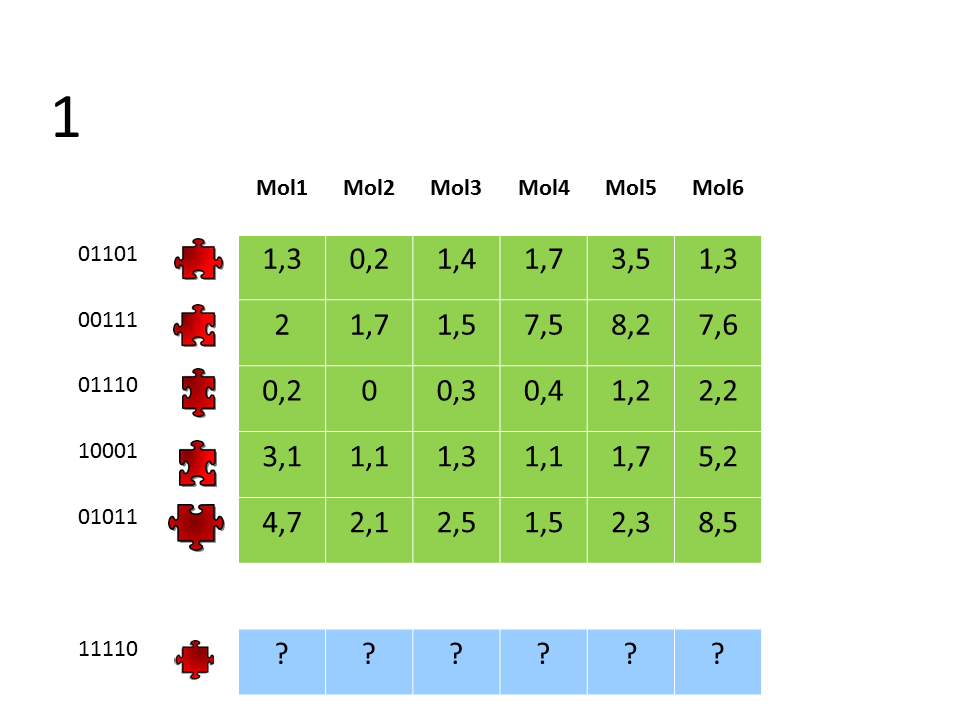} 
 \includegraphics[width=6cm]{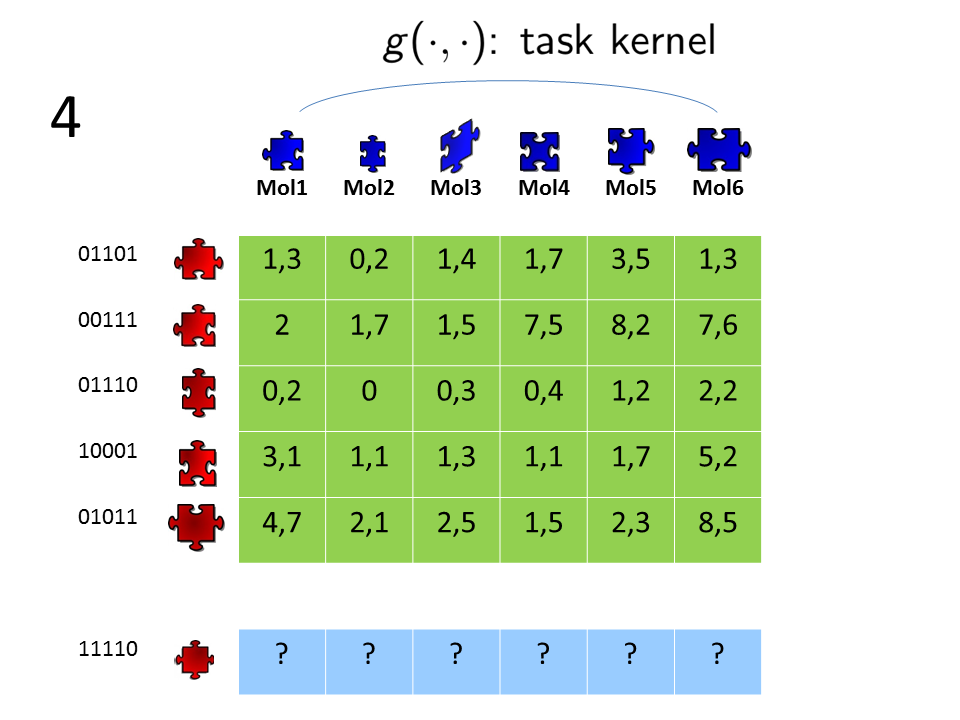} \\
 \includegraphics[width=6cm]{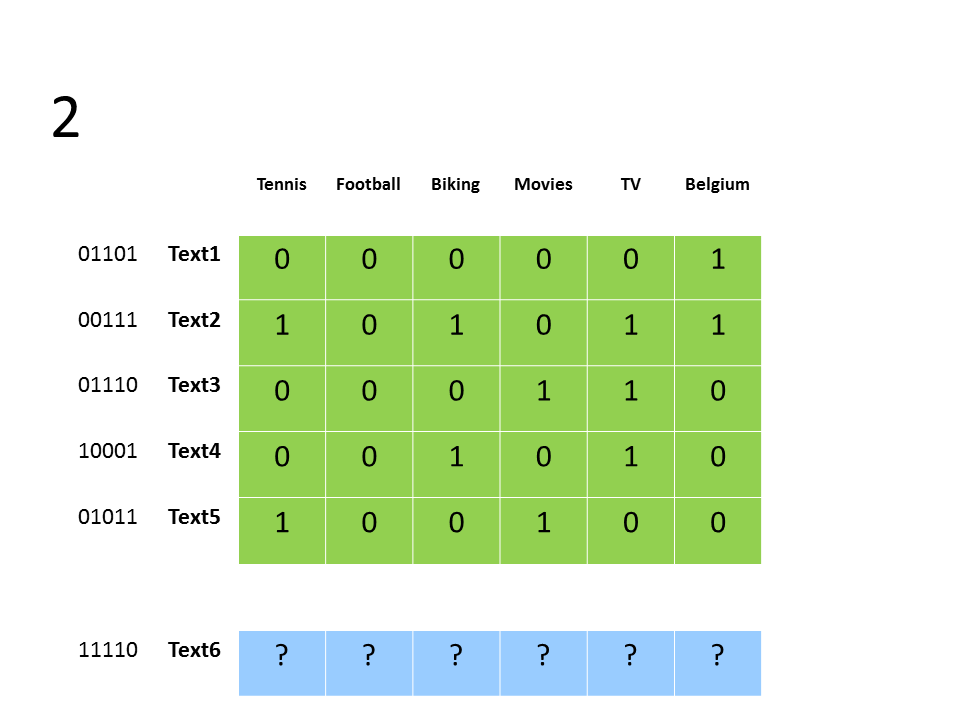}
 \includegraphics[width=6cm]{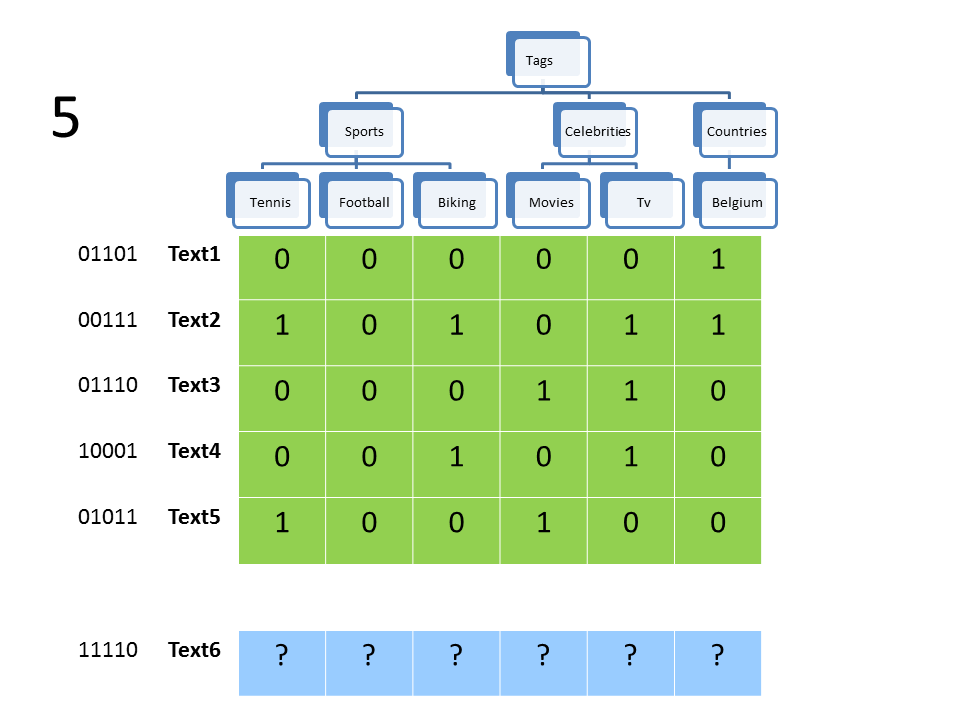} \\
 \includegraphics[width=6cm]{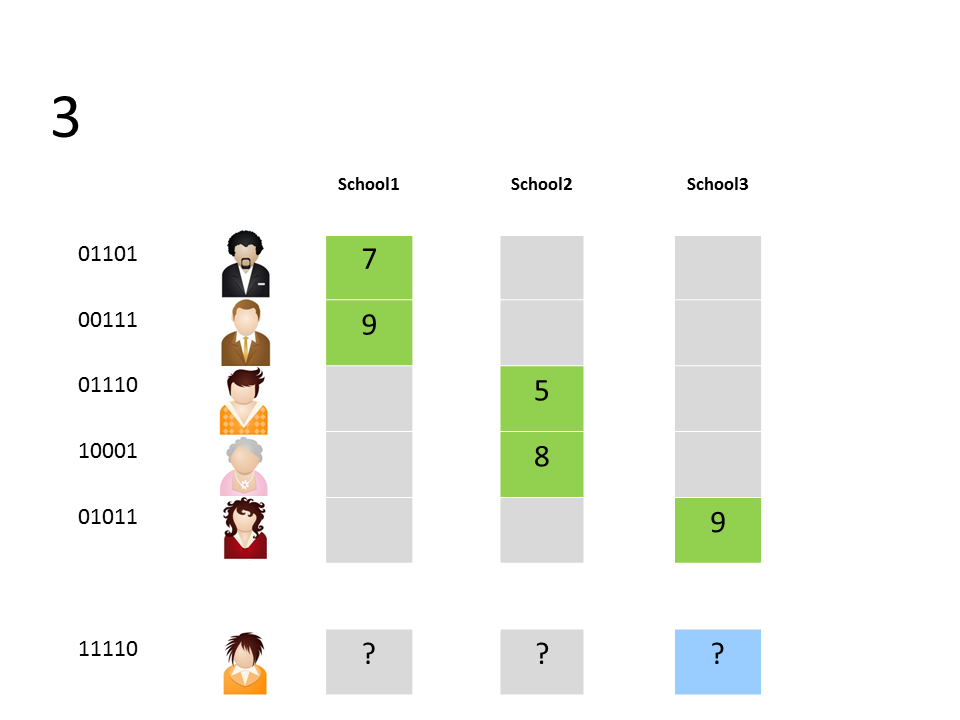}
 \includegraphics[width=6cm]{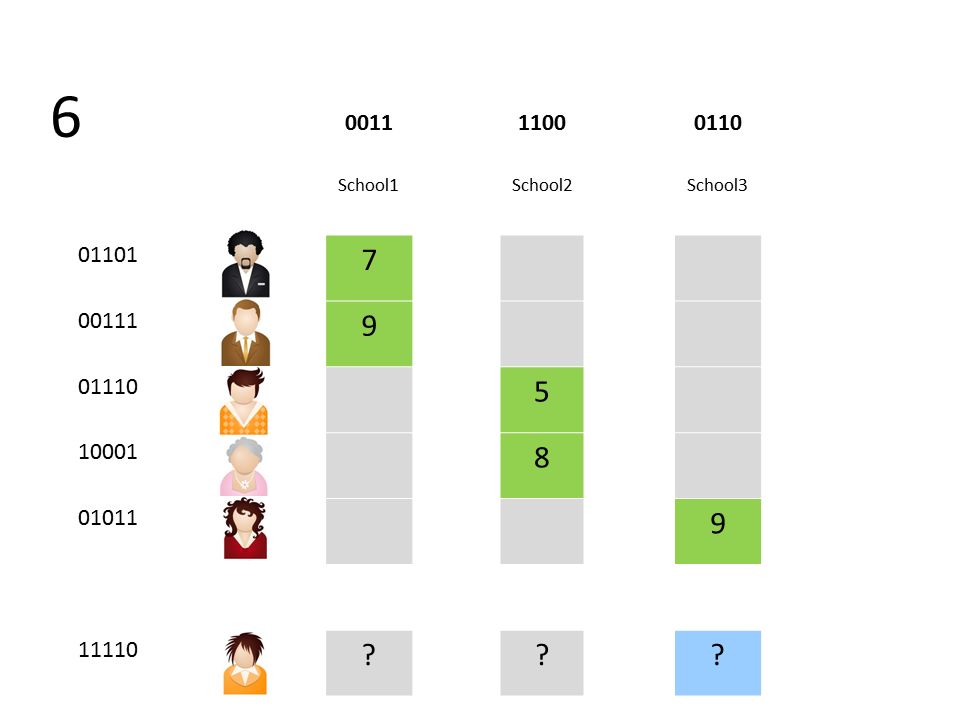}
\end{center}
\caption{A visualization of six prototypical multi-target prediction problems: (1) drug design as an example of a multivariate regression problem, (2) document categorization as an example of a multi-label classification problem, (3) student grading as an example of a multi-task learning problem, (4) drug design with a similarity measure for targets, (5) document categorization with a hierarchy for targets, (6) student grading with target features. See Examples~1 to 6 in the text for more details.}
\label{fig:basicexamples}
\end{figure}

\begin{example}
Consider the problem of predicting whether a protein will bind to a set of experimentally developed small molecules. This is an important application in the area of drug design, where machine learning methods can help in screening interesting novel chemical compounds. Using multi-target prediction methods, high-potential compounds can be selected and analyzed further during (more rigorous) wet-lab experimentation. Figure~\ref{fig:basicexamples}-1 illustrates the data that might be generated in an application of that kind. For a set of training proteins (first five rows, in green), the binding affinity with a set of small molecules (represented as columns) has been verified experimentally. Furthermore, we assume that we have additional information about the proteins available, in the form of a feature representation or a more structured representation such as a graph (shown as red puzzle pieces in the picture). One can use this data to train a multi-target prediction model that is able to predict, on the same set of small molecules, binding affinities for new proteins that have not yet been analyzed in the wet-lab. As experimental verification of binding affinities delivers continuous scores that represent binding strength, one arrives at a multivariate regression problem.    
\end{example}

\begin{example}
In the area of document categorization, assigning appropriate category tags to documents is important but laborious and time consuming. Hence, one often applies multi-target prediction methods to automate this tagging process. As shown in Figure~\ref{fig:basicexamples}-2, one might have asked a human annotator to provide a training dataset with relevant tags being assigned to a set of documents. Using a feature representation of those documents, such as a bag-of-words representation, one can then train a multi-target prediction method that will be able to assign tags to new documents outside the training dataset. More than one tag might be relevant for a particular document, so automated document categorization boils down to solving a multi-label classification problem.     
\end{example}

\begin{example} We consider the application of predicting student marks in the final exam for a typical high-school course, let's say mathematics. Forecasting those marks at an early stage, prior to the start of a course, might be useful to give students advise with respect to study directions and perspectives. Using historical student records that might be stored by schools, one can train a multi-target prediction method that estimates the exam marks of newly entering students. To this end, one would typically construct a dataset that looks like Figure~\ref{fig:basicexamples}-3, where columns represent different courses or different schools. If one aims for estimating the marks for different courses or different schools simultaneously, one ends up with solving a multi-task learning problem.
\end{example}


Examples 1--3 discuss basic multi-target prediction settings in which no additional knowledge about the targets is known. Example~1 is a multivariate regression problem, as one aims to predict the values of continuous variables, whereas Example~2 is a multi-label classification problem with binary target variables. 
Example~3 depicts an application that can only be tackled with multi-task learning methods. As students usually attend only one school, only one label will be observed for per student. The other labels are unknown and typically irrelevant---they are represented by grey cells in Figure~\ref{fig:basicexamples}-3. Likewise, in the prediction phase, it is meaningful to restrict predictions to the school of the student, or at least a subset of schools for which the student is considering an enrolment. In most multi-task learning problems, only one task or a subset of tasks is relevant for a given instance. We therefore introduce the following formal definitions to include multivariate regression, multi-label classification, and multi-task learning in our framework. 

\begin{definition} {\bf (Multivariate regression)}
A multivariate regression problem is a specific instantiation of the general framework, which exhibits the following additional properties: 
\begin{enumerate}
\item[P5.] The cardinality of $\mathcal{T}$ is $m$. This implies that all targets are observed during training. 
\item[P6.] No side information is available for targets. Without loss of generality, we can hence assign the numbers $1$ to $m$ as identifiers to targets, such that the target space is $\mathcal{T} = \{1,...,m\}$. 
\item[P7.] The score matrix $Y$ has no missing values. 
\item[P8.] The score set is $\mathcal{Y} = \mathbb{R}$. 
\end{enumerate}
\end{definition}

\begin{definition} {\bf (Multi-label classification)}
A multi-label classification problem is a specific instantiation of the general framework, which exhibits the following additional properties: 
\begin{enumerate}
\item[P5.] The cardinality of $\mathcal{T}$ is $m$; this implies that all targets are observed during training. 
\item[P6.] No side information is available for targets. Again, without loss of generality, we can hence identify targets with natural numbers, such that the target space is $\mathcal{T} = \{1,...,m\}$. 
\item[P7.] The score matrix $Y$ has no missing values. 
\item[P8*.] The score set is $\mathcal{Y} = \{0,1\}$. 
\end{enumerate}
\end{definition}

\begin{definition} {\bf (Multi-task learning)}
A multi-task learning problem is a specific instantiation of the general framework, which exhibits the following additional properties: 
\begin{enumerate}
\item[P5.] The cardinality of $\mathcal{T}$ is $m$; this implies that all targets are observed during training. 
\item[P6.] No side information is available for targets. Again, the target space can hence be taken as $\mathcal{T} = \{1,...,m\}$.   
\item[P8{**}.] The score set is homogenous across columns of $Y$, e.g., $\mathcal{Y} = \{0,1\}$ or $\mathcal{Y} = \mathbb{R}$.
\end{enumerate}
\end{definition}

In the above definitions, we adopt a matrix view for multi-task learning. This means we consider a matrix of targets, where rows are indexed by the number of training instances, and columns are indexed by the different tasks. A cell $(i,j)$ is only filled with a value $y_{ij}$ if instance $i$ is contained in the training dataset of task $j$. We will use the letter $Y$ to denote the resulting sparsely filled matrix. Figure~1-3 depicts an example of such a matrix. 

From this matrix viewpoint, multivariate regression and multi-label classification arise as special cases (when the matrix $Y$ has no missing values, like in the examples in Figure~\ref{fig:basicexamples}). Multi-task learning methods can usually process multivariate regression and multi-label classification datasets with no problems. In fact, many authors of multi-task learning papers turn out to be analyzing multivariate regression or multi-label classification datasets in their experimental studies, because such datasets are widely available. However, albeit not always visible in such experimental studies, multi-task learning methods are usually more general than multivariate regression and multi-label classification methods.

Multivariate regression and multi-label classification can be further generalized to targets with other types of values, such as nominal, ordinal, or mixed. This setting basically omits Property $8^{**}$. It is sometimes referred to as multi-dimensional classification, even though it has been rarely investigated in the machine learning literature---see e.g.\ \citep{Bielza2011,Read2013}. Another problem that can be seen as an instantiation of our framework is label ranking, where each instance is associated with a ranking (total order) of the targets \citep{Huellermeier_et_al_2008}. Thus, the score $y_{ij} \in \{1,...,m\}$ for a pair $(\bx_i, \bt_j)$ is the position of $\bt$ in the ranking associated with $\bx$, i.e.,  each row of the score matrix $Y$ is a permuation of $\{1,...,m\}$.

\subsection{Problems that involve side information for targets}

Let us now extend multivariate regression, multi-label classification, and multi-task learning to settings where additional side information about the target space is available. To this end, we adjust the three examples from above slightly. 

\begin{example} In the drug design application of Example~1, we treated proteins as instances and small molecules that can potentially bind to those proteins as targets. Let us assume that, in addition to the graph-based representation for the proteins, a representation for the target molecules is also available, as shown in Figure~\ref{fig:basicexamples}-4. This is a common situation in research on protein-ligand prediction, where the representation for the targets is either graph-based or feature-based. The resulting machine learning setting can be interpreted as a dyadic prediction problem. 
\end{example} 

\begin{example}
Let us return to the document categorization application that was described in Example~2. Often document categories are organized in a hierarchy that describes the degree of relatedness among document tags. An example is shown in Figure~\ref{fig:basicexamples}-5. Problems of that kind, where the target space is equipped with a hierarchical structure, are referred to as hierarchical multi-label classification problems. Naturally, state-of-the-art algorithms will try to exploit the structure for better prediction.
\end{example}

\begin{example}
In the student mark forecasting application that was considered before, one might also assume that additional information about the targets is given. One might be able to collect all sorts of variables about schools and courses, such as geographical location, qualifications of the teachers, reputation of the school, etc. Figure~\ref{fig:basicexamples}-6 visualizes such a situation with binary features. One might be able to improve multi-task learning algorithms by taking the features into account. This side information forms a key element for tackling transfer learning and zero-shot learning problems. 
\end{example}

Examples 4-6 further extend basic MTP settings to situations where various types of side information about targets is available. Example~4 illustrates this with the case of a structured representation in the form of a molecular graph. Example~5 assumes a hierarchy that describes relations among targets in a specific manner, and Example~6 considers feature representations for targets. Applications of that kind lead to more complex multi-target prediction problems that are often referred to as dyadic prediction, link prediction, or network inference settings---see e.g. \citep{Menon2010,Schafer2015}. In this area, one can distinguish algorithms that model vector representations or structured target representations, as well as methods that model target relations. This will be further discussed in Section~3.  

Dyadic prediction, link prediction, and network inference are general terms that cover a wide range of problems. Generally speaking, they cover problems that obey the four properties listed in Definition~1. The labels $y_{ij}$ can be arranged in a matrix $Y$, which is often sparsely filled. Thus, one may argue that dyadic prediction is nothing else than multi-task learning with task features. However, the multi-task learning terminology is rarely used in the dyadic prediction literature. Dyadic prediction problems emerge in a variety of application domains, including product recommendation, social network analysis, drug design, various bioinformatics applications, and game playing \citep{Basilico2004,park2009pairwise,Stock2014,Benhur2005,Kashima2009,Pelossof2015,Jacob2008a,pahikkala2010reciprocalkm,pahikkala2013efficientcondrank}. 



\subsection{Inductive versus transductive learning problems}

One may argue that the problems analyzed in Examples~1-3 are inductive w.r.t.\ instances and transductive w.r.t.\ targets. Predictions need be be generated for novel instances, whereas the set of targets is known beforehand and observed during the training phase.   
Side information is of crucial importance for generalizing to novel targets that are unobserved during the training phase, such as a
novel target molecule in the drug design example, a novel tag in the document annotation example, or a novel course in the student grading example. 

\begin{figure}
\includegraphics[width=6cm]{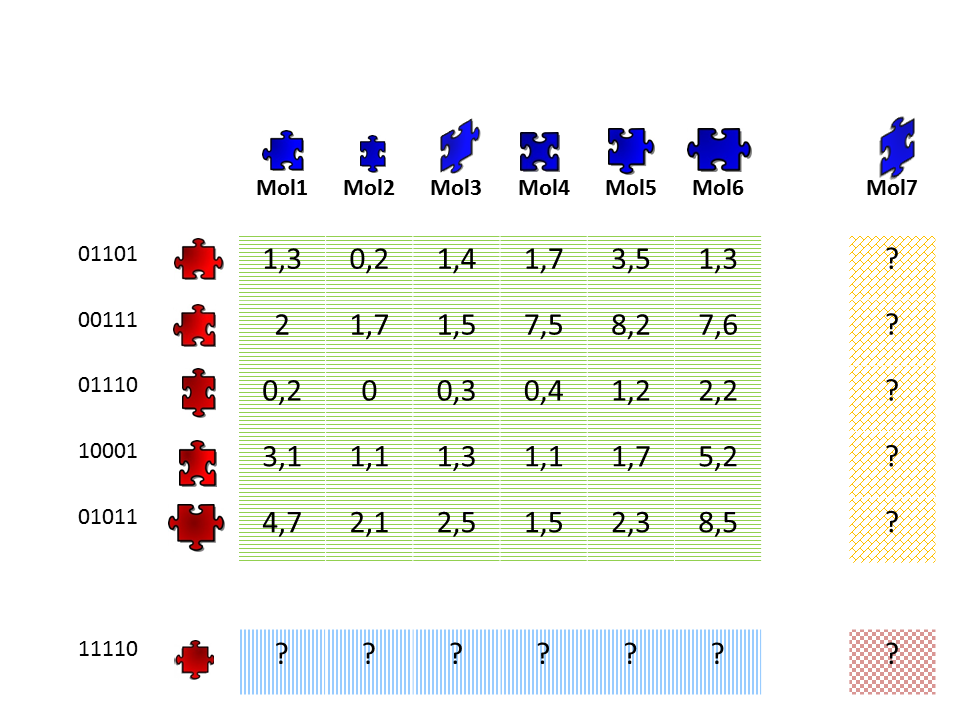} \hspace{0.5cm}
\includegraphics[width=6cm]{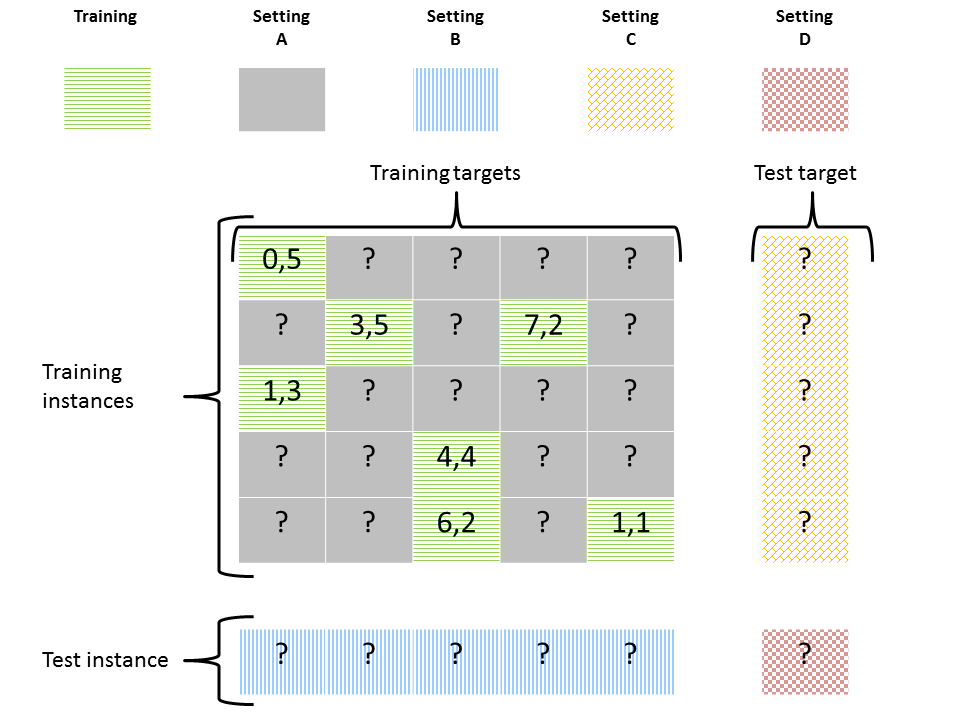}
\caption{Visualization of the different types of prediction problems that arise when prior knowledge of the target space is modelled. Left: an extension of the drug design example, where predictions have to be made for a novel protein (Setting B in blue), a new ligand (Setting C in yellow) or a combination of a  new protein and ligand (Setting D in red). Right: a general overview of the four settings that can be distinguished. Setting A depicts the situation where missing values need to be imputed in the matrix. See text for details.}
\label{fig:foursettings}
\end{figure} 

This is visualized in Figure~\ref{fig:foursettings}, where the red part of the dataset alludes to a protein-drug pair that was not observed during the training phase. In the multi-task and transfer learning literature, this type of problem setting is known as zero-shot learning \citep{Larochelle2008}. Since no training data is available for the novel target, the problem is intrinsically more difficult than those analyzed in Section~2.2 \citep{park2012flaws}. In this work, we define zero-shot learning as follows.

\begin{definition} {\bf (Zero-shot learning)}
A zero-shot learning problem is a specific instantiation of the general framework with the following additional property: 
\begin{enumerate}
\item[P5*.] The cardinality $m^*$ of the set $\mathcal{T}$ is bigger than $m$. Some targets are hence not observed during training, but may nevertheless appear at prediction time.  
\end{enumerate}
\end{definition}

By substituting P5 with P5*, one now tackles problems that are inductive instead of transductive w.r.t.\ targets.  The same subdivision can be made for instances. As pointed out by \citet{Pahikkala2014}, in total four different settings can be distinguished in the presence of side information about instances and targets. In Figure~\ref{fig:foursettings}, Setting D resembles the zero-shot learning setting, whereas Setting B coincides with the problems that we discussed in Section~2.2. In addition, one can also think of Setting C (predicting scores for a novel target on an instance that was contained in the training dataset) and Setting A (predicting scores for an instance-target combination, where the target and the instance were already seen during the training phase).  Combinations of those settings also exist. For example, one can distinguish zero-shot learning settings where only targets that are unknown during training need to be predicted in the test phase (thus Setting D), versus settings where both training and test targets have to be predicted for novel instances (a combination of Setting B and D). The latter, for example, happens in image classification tasks, where it is referred to as generalized zero-shot learning \citep{Rohrbach2011a,Xian2017}. Setting C is inductive w.r.t.\ targets and transductive w.r.t.\ instances. Setting A is transductive w.r.t.\ instances and targets. 

Setting C is in theory identical to Setting B. When side information is available for both instances and targets, the two settings become interchangeable. In Example~4, we referred to proteins as instances and small molecules as targets, but in principle, one could transpose the matrix without changing the algorithms. However, interchanging the rows and columns of a particular dataset and applying a particular multi-target prediction algorithm on the two scenarios will in practice lead to different results. Often, one of the two sources of side information is more informative than the other, and many algorithms take this implicitly into account. Take the document categorization example, for which textual descriptors on the document level are much more informative and richer than the hierarchical structure on the tag level. Therefore, modelling the side information on the document level is of uttermost importance, while a reasonable performance might still be expected even if the side information on the tag level is ignored.      

In contrast, Setting A results in yet a different type of problem, namely that of matrix completion. Both the targets and the instances are observed, albeit not for all instance-target combinations. In principle, one can solve this setting with methods that ignore side information of the target and the instance space. In such cases, latent representations of instance and target space are both deduced using only the instance-target interaction matrix. In light of Examples~1-3, this means that the feature representations of proteins, documents, and students are not used. In fact, they do not even have to be available. As a result, we formally define matrix completion problems as follows.  

\begin{definition} {\bf (Matrix completion)}
A matrix completion problem is a specific instantiation of the general framework with the following additional properties: 
\begin{enumerate}
\item[P5.] The cardinality of $\mathcal{T}$ is $m$. This implies that all targets are observed during training. 
\item[P6.] No side information is available for targets. Without loss of generality, we can hence assign identifiers to targets from the set $\{1,...,m\}$ such that the target space is $\mathcal{T} = \{1,...,m\}$.
\item[P9.] The cardinality of $\mathcal{X}$ is $n$. This implies that all instances are observed during training. 
\item[P10.] No side information is available for instances. Without loss of generality, we can hence assign identifiers to instances from the set $\{1,...,n\}$, such that the instance space is $\mathcal{X} = \{1,...,n\}$.
\end{enumerate}
\end{definition}

Matrix completion methods are extensively used in areas such as recommender systems, social network analysis, and biological network inference. They are therefore also known as collaborative filtering and link prediction methods. One can distinguish different versions of matrix completion. One may consider matrices that contain continuous, ordinal or binary values, but situations with presence-only data are also common. Observed entries in the matrix then correspond to known positive interactions between the items in the rows and the columns, and the goal consists of predicting more positive interactions from the large pool of missing entries. The assumption made in such cases is that the overwhelming majority of the missing entries correspond to negative interactions, with the exception of a few positive ones that need to be found. Properties 5 and 9 characterize the transductiveness of matrix completion methods. Properties 6 and 10 characterize the difference between a pure matrix completion setting, and a hybrid setting that simultaneously exploits the structure of the matrix and side information to generate predictions.   

\subsection{Beyond MTP: Problems that are not covered}

Our formal framework is rather generic and covers several machine learning problems as special cases, including those discussed above. In principle, every prediction problem with (original) output space $\mathcal{Y}$ could be seen as a special case, simply by taking $\mathcal{T} = \mathcal{Y}$ as the set of tasks and $\{0,1\}$ as a score set. This amounts to treating each candidate output as a target, and reinterpreting the task to predict an output from $\mathcal{Y}$ as predicting, for each candidate value, whether or not the sought output corresponds to that candidate. Consequently, a consistent prediction has to obey strong (deterministic) dependencies between the targets: It should assign the score 1 to exactly one target and 0 to all the others.  As two examples, consider multi-class classification and structured output prediction (SOP), which can be formalized, respectively, as specializations of multi-label classification and zero-shot learning. 

\begin{definition} {\bf (Multi-class classification)}
A multi-class classification problem is a specific instantiation of the general framework with the following additional properties: 
\begin{enumerate}
\item[P5.] The cardinality of $\mathcal{T}$ is $m$. This implies that all targets are observed during training. 
\item[P6.] No side information is available for targets. Without loss of generality, we can hence assign identifiers to targets from the set $\{1,...,m\}$ such that the target space is $\mathcal{T} = \{1,...,m\}$. 
\item[P7.] The score matrix $Y$ has no missing values. 
\item[P8*.] The score set is $\mathcal{Y} = \{0,1\}$. 
\item[P11.] Each row in $Y$ has a single ``positive" entry, and all other entries are zero. 
\end{enumerate}
\end{definition}

\begin{definition} {\bf (Structured output prediction)}
A structured output prediction problem is a specific instantiation of the general framework with the following additional properties: 
\begin{enumerate}
\item[P5*.] The cardinality $m^*$ of the set $\mathcal{T}$ is bigger than $m$. Some targets are hence not observed during training, but they appear at test time.  
\item[P7.] The score matrix $Y$ has no missing values. 
\item[P8*.] The score set is $\mathcal{Y} = \{0,1\}$. 
\item[P11.] Each row in $Y$ has a single ``positive" entry, and all other entries are zero. 
\end{enumerate}
\end{definition}


Note that these encodings of the problems correspond to what is known as a 1-versus-rest decomposition, a specific reduction technique that makes prediction problems with $|\mathcal{Y}| > 2$ amenable to binary classification. Obviously, other reduction techniques such as all-pairs or, more generally, error correcting output codes (ECOC) lead to similar representations: The problem of predicting the output for an instance $\bx \in \mathcal{X}$ is decomposed into a set of binary prediction problems, and each of these problems is seen as a task. 



While formally possible, we are not in favor of considering problems such as multi-class classification and structured output prediction as special cases of MTP. The reasons include both conceptual and algorithmic aspects. Conceptually, the view of each candidate prediction as a separate target appears to be rather artificial. Actually, one is still interested in a single prediction, not multiple ones. To comply with the corresponding consistency constraints, a kind of post-processing (like the decoding step in ECOC) is normally required. Algorithmically, it is also clear that the multi-target perspective is not typical of methods for SOP. Instead, such methods are specifically tailored for output spaces in SOP, which are often huge but equipped with a strong structure. 



In this paper, we will also exclude other (multi-target) prediction problems where the ground truth cannot be represented in a matrix format with optional side information for rows and columns. This includes problems that involve multi-instance learning representations \citep{Zhou2007} and dyadic feature representations \citep{Vanpeer2017}. The latter scenario occurs, for example, in bio-informatics and recommender systems applications, when features are available that describe an interaction between an instance and a target, e.g., the day of the week when a user clicked a specific item on a website.

\section{A unifying view on MTP methods}


In this section, we present a unifying view on MTP methods. To this end, we categorize methods according to some of the properties that were discussed in Section~2.  In particular, we believe that the properties characterizing Settings A, B, C and D are specifically important in this regard, and suitable to distinguish MTP methods. We therefore put a main emphasis on those properties, while some other properties of Section~2 are not further considered. As a result, our categorization of MTP methods does not yield a one-to-one mapping with MTP problems. Table~\ref{tab:sections} gives an overview of the subdivision we propose. Most of the methods to be discussed can be applied to Settings B and C. Some of them are also applicable to Setting D, while Setting A normally calls for more specifilized methods.  


\begin{large}

\begin{center}
\begin{table}[h]
\caption{Categorization of MTP methods.}
\label{tab:sections}
\begin{tabular}{ll}
\hline
Group of methods & Applicable setting \\
\hline
\hline
Section \ref{sec:methods-that-do-not-use-domain-knowledge}: Similarity-enforcing methods & B and C   \\ 
Section \ref{sec:methods-that-use-target-relations}: Relation-exploiting methods & B, C and D  \\
Section \ref{sec:methods-that-learn-target-relations}: Relation-constructing methods & B and C \\
Section \ref{sec:methods-that-use-target-representations}: Representation-exploiting methods & B, C and D \\
Section \ref{sec:methods-that-learn-target-representations}: Representation-constructing methods & B and C \\
Section \ref{sec:matrix-completion-and-hybrid-methods}: Matrix completion and hybrid methods & A \\
\hline  
\end{tabular}
\end{table}
\end{center}
\end{large}

In what follows, we focus on the average squared prediction error over all targets as a loss function to be minimized. Formally, for a given pair $(\bx, \bt)$, a predictor $f$ yields a prediction $\hat{y}= f(\bx, \bt)$ of an actual outcome (score) $y$. Let $\by$ be a vector of actual output values of size $m$ (or $m^*$ in the case of zero-shot learning), and let $\hat{\by}$ be a vector of the corresponding predictions. Then, the squared-error loss is given by
\begin{equation}
L(\by,\hat{\by}) = \sum_{i=1}^{m} (y_i - \hat{y}_i)^2 \,.
\label{eqn:squared-error-loss}
\end{equation}
In spite of obvious shortcomings, both in regression and classification settings, this loss is somewhat representative in the sense of being used in the majority of algorithms for multivariate regression and multi-task learning, and therefore suitable for explaining general concepts. It has the additional advantage of being applicable  for both numerical and binary target variables, again making it a natural choice for the unifying view that we intend to provide. All the methods presented below can be tailored for minimizing this loss function.  

In multi-label classification, many methods optimize loss functions that are more complex and not decomposable over targets. Because those methods only appear in the MLC literature, an in-depth discussion is beyond the scope of this paper. Instead, we refer the interested reader to \citet{Demb2012a}. 

We start our discussion on multi-target methods by introducing a simple baseline that does not exploit any target dependencies. Subsequently, we present more and more complex models that tackle the problem of multi-target prediction in different manners. 

\subsection{Independent models}
\label{sec:independent-models}

The most straight-forward approach to solving MTP problems consists of constructing one model $f_i$ for every target independently, and to concatenate the predictions of these models into the sought multi-target prediction. 
In the multi-label classification community, this approach has an own name and is known as \emph{binary relevance learning}~\citep{Tsoumakas_and_Katalos_2007}. 
In order to introduce the approach more formally, let $f_i(\vec{x})$ denote the score (continuous or discrete) assigned to the $i$th target for instance $\vec{x}$. Using linear basis function models, the model for the $i$th target can be represented as 
\begin{equation}
f_i(\vec{x}) = \vec{a}_i^\intercal \phi(\vec{x}) \,,
\label{eq:binrel}
\end{equation}
where $\vec{a}_i$ is a target-specific parameter vector and $\phi(\vec{x})$ a feature representation that is either given explicitly or first obtained from another representation in a preprocessing step.  Multivariate ridge regression is probably the most basic model of this kind. This approach fits a linear model in a regularized least-squares fashion, while ignoring potential dependencies among targets \citep{Hastie_et_al_2007}. More formally, multivariate ridge regression boils down to minimizing the following objective function:  
\begin{equation}
\label{eq:multiridge}
\min_A ||Y - XA ||^2_F +  \sum_{i=1}^m \lambda_i \, ||\vec{a}_i||^2 \,,
\end{equation}
where $Y$ is the $n \times m$ target matrix with $n$ the number of observations and $m$ the number of targets. Denoting the number of features by $p$, $X$ and $A$ are $n \times p$ and $p \times m$ matrices, respectively, which are constructed as follows:
\begin{equation}
\label{eq:notation}
X = \begin{bmatrix} \phi(\vec{x}_1)^T \\ \vdots \\ \phi(\vec{x}_n)^T \end{bmatrix} \qquad A = [\vec{a}_1 \quad \cdots \quad \vec{a}_m] \,.
\end{equation}
$||.||_F$ is the Frobenius-norm, so that the left-hand side corresponds to the sum of squared error losses (\ref{eqn:squared-error-loss}) on the $m$ targets. The $\lambda_i$ are regularization parameters. 

Optimization problem (2) can be solved by matrix inversion in a least-squares algebraic fashion. This is efficient because the  inversion needs to be done only once for all targets. The formulation assumes a linear statistical model, but independent models can be constructed using any well-known method for regression (in the case of continuous variables as targets) or classification (in the case of binary variables as targets). 

Can one perform better with more sophisticated approaches that exploit dependencies among targets? Unsurprisingly, the answer is affirmative. Many of the approaches that were introduced over the last decade in each of the MTP subcommunities are able to outperform the simple baseline that treats every target as an independent problem. Generally speaking, these methods put soft constraints on the values that targets can take, so they assume that not all combinations of values are equally likely to occur: some combinations of target values have a higher probability of occurrence than others, yielding a dependency among targets in a statistical sense.


Authors introducing a novel multi-target prediction method often emphasize the exploitation of target dependencies as the key to improved performance. Without questioning explanations of that kind, we like to note that improvements over independent predictions can be achieved even for problems without any statistical dependence between the targets. This insight is illustrated, for example, by the famous James-Stein paradox, showing that, under certain conditions, maximum likelihood estimation is not optimal (or, using statistical terminology, not admissible) for estimating the mean of a multivariate Gaussian \citep{James1961}. Since the James-Stein estimation principle is less known among machine learning scholars, we provide a short description in Appendix~A.

\subsection{Similarity-enforcing methods}
\label{sec:methods-that-do-not-use-domain-knowledge}

We start our overview of MTP methods by explaining a few simple methods that do not consider any particular domain knowledge about targets. Such methods can be found in multivariate regression, multi-label classification, and multi-task learning, and they are usually only applicable to Settings B and C.  The central idea in those simple methods is that models for different targets should behave \emph{similar}. This can be achieved, for example, by enforcing similarity of the parameterizations of models for different targets. To this end, one usually defines a mathematical objective that incorporates a specific regularizer. In early work on multi-task learning, \citet{Evgeniou_2004}, \citet{Evgeniou2005}, and \citet{Jalali2010} solve the following optimization problem: 
\begin{equation}
\label{eq:meanreg}
\min_A ||Y - XA ||^2_F + \lambda \sum_{i=1}^m ||\vec{a}_i - \frac{1}{m} \sum_{j=1}^m \vec{a}_j||^2 \, ,
\end{equation}
with $A$ the parameter matrix, $X$ the feature matrix, and $Y$ the target matrix, as defined before. 
In this way, one penalizes for deviations from the ``mean target". The parameter estimates are biased to the mean, in an attempt to reduce the variance and prevent overfitting. 

Another common way of obtaining restricted models with less flexibility (compared to modelling each target independently) consists of implementing feature selection strategies that retain the same features for different targets. Joint feature selection leads to a model with less parameters compared to binary relevance, and this may in turn result in performance gains. The idea has been particularly popular in the area of multi-label classification \citep{Obzinski2010,Gu:2011,Kong2012,Spolaor:2016}. Even though most of those authors put the emphasis on computational efficiency as a motivation, selecting a subset of identical features for different targets can also improve the predictive performance. Similar ideas are appearing in the literature on multi-task learning \citep{Zhou2011} and in joint regression of binary and continuous response variables \citep{Zhang2012}. An embedded feature selection strategy can be obtained, for example, by a combination of L1 and L2 norms:
$$
\min_A ||Y - XA ||^2_F + \lambda \sum_{j=1}^p ||\vec{a}_j||^2  
$$ 
In contrast to (\ref{eq:meanreg}), the sum is taken over the features. Unlike (\ref{eq:notation}), the vectors $\vec{a}_j$ now represent the rows of matrix $A$:
$$A = \begin{bmatrix} \vec{a}_1^T \\ \vdots \\ \vec{a}_p^T \end{bmatrix}$$
Thus, $\vec{a}_j$ specifies the set of parameters for the $j$th feature, across different targets. The combination of the two types of norms results in group sparsity, so that a set of identical features is selected for different targets. 

Yet another straight-forward approach to exploiting dependencies among targets is stacking \citep{Wolpert1992}. This approach was initially introduced as an ensemble learning technique in conventional classification and regression settings. Later on, it has been extended for multi-label classification  \citep{Godbole_Sarawagi_2004,Cheng_Hullermeier_2009}. In the statistical literature, a slightly more advanced method has shown to improve the predictions for multivariate regression \citep{Breiman_Friedman_1997}. Although one could  easily think of extensions for multi-task learning problems, too, the method has been less popular in that area.   

Stacking implements a two-step procedure. In the first step, a model is fitted for every target individually, as discussed in Section~\ref{sec:independent-models}. Then, in a second step, the predictions obtained by each model are used as a feature representation to train a second series of models, again one for each target. Obviously, these second-order models are able to capture dependencies between different targets, as they seek to represent the prediction of one target as a function of the (predictions of) the others. Thus, the goal of the second step is to discover dependencies among targets, and to identify, for each target in turn, a set of other targets that help improve predictions. More formally, with the shorthand-notation 

$$\vec{f}(\vec{x}) = \begin{bmatrix} f_1(\vec{x}) \\ f_2 (\vec{x}) \\ \vdots \\ f_m(\vec{x}) \end{bmatrix}
$$ 
for the independent models of (\ref{eq:binrel}), the general scheme  of stacking can be expressed as follows:
\begin{equation}
\label{eqn:slp_1}
\vec{f}^*(\vec{x}) = \mathbf{b} (\mathbf{f}(\vec{x}), \vec{x})\, ,
\end{equation}
with $\vec{f}^*(\cdot)$ the predictions made by stacking and $\mathbf{b}(\cdot)$ a second-stage model that shrinks or regularizes the solution of the initial models. \citet{Breiman_Friedman_1997} show formally that stacking prevents overfitting. As soon as some regularization is present in the second-order models, one introduces a bias by encouraging models for different targets to learn similar parameters. The same authors also reveal some connections with James-Stein estimation. 

Two appealing properties of stacking are its ease of implementation and its generality, as it can be applied in tandem with any type of classifier or regressor. In addition, the second-level model $\mathbf{b}$ can be trained on the first-level predictions $\mathbf{h}(\vec{x})$ alone or in concatenation with the original features $\vec{x}$. In the case of classification, there is also a choice between feeding the second-level model with binary predictions or with continuous scores obtained from scoring functions or probabilities, if such outputs are delivered by the classifier. 

The idea of enforcing that models for different targets behave similarly can also be found in neural networks, starting from classical papers such as \citep{Caruana1997}, till modern deep learning architectures, such as deep convolutional nets for multi-label classification \citep{Wei2016}. Such methods are based on weight sharing in the final network layer, which usually includes $m$ nodes for a multi-target prediction setting with $m$ targets. Likewise, the common idea of pre-training a deep neural network on an auxiliary task with a lot of training data, and further optimizing it for another task with limited training data, can be interpreted as weight sharing---see e.g.\ \citep{Donahue2014,Girshick2014,Oquab2014,Razavian2014,Sermanet2014,Gong2014b}. 


\subsection{Relation-exploiting methods}
\label{sec:methods-that-use-target-relations}

In this section, we discuss MTP methods using side information in the form of target relations, such as hierarchies, graph structures, decision rules, and correlation matrices. This side information might be useful for at least three purposes. First, it might help obtain performance gains for Settings B and C, compared to the methods that were discussed in Section~\ref{sec:methods-that-do-not-use-domain-knowledge}. Second, it can be considered as a key element when generalizing to novel targets in Setting D. Third, it may also effect the computational scalability of algorithms in a positive way. 

When speaking about target relations, it is important to make a distinction between deterministic relations, which are guaranteed to hold for each observation, and probabilistic relations, which only give an indication of the likelihood of occurrence. In multi-label classification, for example, one could observe deterministic relations such as implications, subsumptions, and mutual exclusions between relevant labels of an instance. In label ranking, the values of target variables are positions in a ranking, which means that all target values must be different (each position can only be occupied once). In multivariate regression, targets may represent consecutive time steps, for which constraints such as smoothness or monotonicity over consecutive values might apply. 


For MLC problems such as Example 5 in Section~2, relationships between labels are often represented in the form of a tree or hierarchy, resulting in a set of problems that is known as hierarchical multi-label classification. Many authors have shown that such a hierarchy can be integrated in specific methods \citep{Rousu2006,Barutcuoglu2006,Vens_et_al_2008,Silla2010,Gopal2012,Nam2015}. The hierarchy not only helps to improve predictions, but also allows for defining extended prediction settings that are characterized by specific performance measures. For example, when a classifier is unsure about certain classes, it could be allowed to return intermediate nodes from the tree as prediction instead of the leaf nodes that correspond to single labels \citep{Bi2012}.   

A few authors have presented more general approaches for incorporating deterministic relations that cannot be represented as a hierarchy. \citet{Gopal2013} and \citet{Deng2014} both consider graph structures.  When targets follow a chronological order, chain-based graph structures might be a natural choice. For example, \citet{Zhou:2012} study disease progression prediction as a multi-task learning problem, where tasks follow a chronological order, leading to a formulation where a specific temporal regularizer in the optimization problem is introduced. This regularizer can be seen as a special case of the graph-based regularization of \citet{Gopal2013}: 
\begin{equation*}
\min_A ||Y - XA ||^2_F + \lambda \sum_{i=1}^m \sum_{j \in \mathcal{N}(i)} ||\vec{a}_i - \ \vec{a}_j||^2 \, ,
\end{equation*}
where $\mathcal{N}(i)$ denotes the set of targets that are related to the $i$th target. This formulation shows that methods using target relations are actually very related to those exploiting similarity (cf.\ Section~\ref{sec:methods-that-do-not-use-domain-knowledge}). The available side information essentially allows one to define prior knowledge concerning statistical dependencies that are likely to occur.  

In addition to deterministic relations, there are also methods that incorporate probabilistic relations. In the context of kernel methods, it is natural to represent target similarities by means of so-called output kernels, which can be interpreted as expressing some sort of correlation between targets. Output kernels have been mainly developed for multivariate regression, under the framework of vector-valued kernel functions \citep{Caponnetto2008,baldassarre2012multioutput,Alvarez2012}. Similar kernel-based formulations have been proposed for multi-label classification \citep{Hariharan_et_al_2010}. 

All the above methods essentially assume that the provided information about the target space results in a good representation for modelling target dependencies. This is clearly a necessary condition for improvement---a few authors have also reported performance drops in certain applications, when the given target relations lead to an incorrect representation of target dependencies. 


\subsection{Relation-constructing methods}
\label{sec:methods-that-learn-target-relations}

When target relations to be used as side information are not available in the form of prior knowledge, one can try to construct such relations from the training data. As an important consequence of this approach, note that it excludes a generalization to novel targets. Thus, unlike the methods discussed in the previous section, methods that learn target relations are not applicable to Setting D; instead, they are mainly useful for settings B and C. 

Learning target relations can be useful for several reasons. First of all, domain knowledge is often not available, so that target relations cannot be modelled as prior information. Second, even if such knowledge can in principle be provided, target relations specified by domain experts might be incorrect and hence misleading for machine learning algorithms. A third motivation is computational efficiency, since learning target relations paves the way for methods that scale sublinearly in the number of targets. 

Like in the case where target relations are given instead of being learned, one can distinguish methods w.r.t.\ the type of relations: hierarchies, more general graph structures, and correlation matrices. In hierarchical multi-label classification, where the correctness of a hierarchy could be questioned, one can find several methods that adjust the existing hierarchy in the course of the learning procedure. In that sense, one can distinguish level-flattening algorithms, node removal algorithms, hierarchy modification algorithms, and hierarchy generation algorithms; see \citep{Rangwala2017} for an overview. In those methods, the focus is on improving the predictive performance. Conversely, in extreme multi-label classification, i.e., problems that involve thousands or millions of labels, hierarchies are often learned to obtain a training or prediction time that is logarithmic in the number of labels \citep{Agrawal2013,Weston2013,Prabhu2014,Demb2016}.
 
Graphs more general than hierarchies might be considered. This idea is popular when applying graphical models in multi-target prediction settings, see e.g.\ \citep{Guo2011,Papagiannopoulou2015}. Using directed graphs, one can infer asymmetric relationships among targets \citep{Lee2016}, while similar results might be obtained with specific rule-based systems  \citep{Park2008,Loza2016}. Let us remark that, for many of those methods, it becomes more difficult to say whether the inferred target relations are deterministic or probabilistic, because the learning process usually contains both deterministic and probabilistic elements.  

In some applications, one may cluster targets into a number of non-overlapping groups, such that interactions between targets need to be considered only inside the same group. This can be established by using a cluster-norm in the optimization \citep{jacob2008,Wang2009}. Similarly, clustered multi-task learning was considered in a Bayesian setting by \citet{Bakker2003} via a mixture of Gaussians instead of a single Gaussian prior, and by \citet{Xue2007} via a Dirichlet process prior. Furthermore, \citet{Gong2012} identify outlier targets, resulting in a structure where only information is shared among the inliers. Similarly, clustering methods have been considered in multi-label classification, where both performance gains in prediction and computational complexity have been reported.   

Finally, one can also learn target relations in the form of similarity or correlation matrices. Unsurprisingly, this idea has been very popular in the field of kernel methods. 
The older methods rather assume that the output kernel is given as domain knowledge, whereas more recent methods  try to discover target similarities by learning the output kernel in a convex formulation \citep{Dinuzzo2011,Dinuzzo2013,Jawanpuria2015}. Similarly, one can learn the covariance function among targets in a Bayesian treatment, by assuming a multivariate Gaussian distribution as prior \citep{Zhang2010}.

\subsection{Representation-exploiting methods}
\label{sec:methods-that-use-target-representations}

Instead of using relations between targets, one can also consider to use target representations as side information. Examples~4 and 6 in Section~2 describe two applications where this type of side information is available in the form of a graph representation and a vector representation, respectively. Just like for target relations, this side information can be useful for several purposes. It might boost the predictive performance in Settings B and C, and is essential for generalizing to novel targets in Setting~D.

Compared to target relations, target representations still require the learning algorithm to discover which elements of the representations are useful and which are not. For instance, in the student grading application of Example~6, one still has to figure out which of the school features are useful. Likewise, in the drug design application of Example~4, one would need specialized methods that can handle structured data such as molecular graphs. In contrast, when working with target relations, the side information is usually available in a more direct and a less noisy format.


Historically, kernel methods have played an important role in modelling target representations. Such methods can easily process vectorial representations, as well as structured data. Moreover, they also establish a clear connection with methods that model target relations: a transformation from a primal formulation to a dual formulation in fact implies a transformation from a target representation to a target relation. In the primal formulation,  information about the instance space and the target space is combined by means of a joint feature representation, whereas in the dual a joint kernel is modelled. Such representations typically yield prediction models of the following form:

\begin{equation}
\label{eq:pairwise}
f(\vec{x},\vec{t}) = \vec{w}^T \Psi(\vec{x},\vec{t}) = \sum_{(\bar{\vec{x}},\bar{\vec{t}}) \in \mathcal{D}} \alpha_{(\bar{\vec{x}},\bar{\vec{t}})} \Gamma((\vec{x},\vec{t}),(\bar{\vec{x}},\bar{\vec{t}})) \, ,
\end{equation}
where $\mathcal{D}$ represents the training data as defined before, $\vec{w}$ and $\alpha_{(\bar{\vec{x}},\bar{\vec{t}})}$ are primal and dual parameters, $\Psi(\vec{x},\vec{t})$ is a joint feature representation in the primal formulation, and 
$$\Gamma((\vec{x},\vec{t}),(\bar{\vec{x}},\bar{\vec{t}})) = \Psi(\vec{x},\vec{t})^T \Psi(\bar{\vec{x}},\bar{\vec{t}})$$
is a joint kernel in the dual formulation, with $\vec{t}$ and $\bar{\vec{t}}$ representations of two targets.

The most commonly used pairwise kernel is the Kronecker product pairwise kernel \citep{Basilico2004,oyama2004using,Benhur2005,park2009pairwise,Hayashi2012,Bonilla2007,pahikkala2013efficientcondrank}, resulting in the following mathematical model: 
$$\Psi(\vec{x},\vec{t}) = \phi(\vec{x}) \otimes \psi(\vec{t}) \,, \qquad \Gamma((\vec{x},\vec{t}),(\bar{\vec{x}},\bar{\vec{t}})) = k(\vec{x},\bar{\vec{x}}) \cdot g(\vec{t},\bar{\vec{t}}) \,,$$
with $\phi$ and $\psi$ feature maps in primal form, and $k(\vec{x},\bar{\vec{x}})$ and $g(\vec{t},\bar{\vec{t}})$ any kernel on the instance and target space in dual form. For example, for the drug design application of Example~4, one would typically develop domain-specific kernels for both $k(\vec{x},\bar{\vec{x}})$ and $g(\vec{t},\bar{\vec{t}})$. In the student grading application of Example~6, one would more likely apply conventional kernels for vectors on the given feature representations. For the document categorization application of Example~5, one would perhaps prefer to use other methods from the domain of hierarchical multi-label classification, but one could also apply a kernel that mimics the hierarchical structure, such as the shortest path kernel. 

As an aggregation of the instance kernel $k(\vec{x},\bar{\vec{x}})$ and the target kernel $g(\vec{t},\bar{\vec{t}})$, the Kronecker product pairwise kernel might be a reasonable first choice in experimental studies, as it is known to exhibit universal approximation properties \citep{Stock2016}. Variants also exist for modelling additional domain knowledge concerning relationships between instances and targets \citep{Vert2007,pahikkala2010reciprocalkm,Waegeman2012,pahikkala2013efficientcondrank}). Pairwise kernels are much more general building blocks than the output kernels that were discussed in Sections~\ref{sec:methods-that-use-target-relations} and~\ref{sec:methods-that-learn-target-relations}. Pairwise kernels model a similarity score based on target representations, whereas no target representations are available for output kernels.  


Many pairwise kernels can be easily plugged into standard kernel methods for binary classification or regression, such as support vector machines or kernel ridge regression. Approaches of that kind are often critisized for computational reasons, and claimed to be infeasible for large sample studies, but efficient implementations based on algebraic shortcuts exist \citep{VanLoan2000,Kashima2009,Raymond2010scalable,Alvarez2012,pahikkala2013efficientcondrank}.  More recently, \citet{Pahikkala2014} and \citet{Romera-paredes2015} independently proposed algorithms that avoid the explicit construction of pairwise kernels. Those algorithms have computational advantages w.r.t.\ cross-validation and online training.  

\begin{figure}
\begin{center}
\includegraphics[scale=0.4]{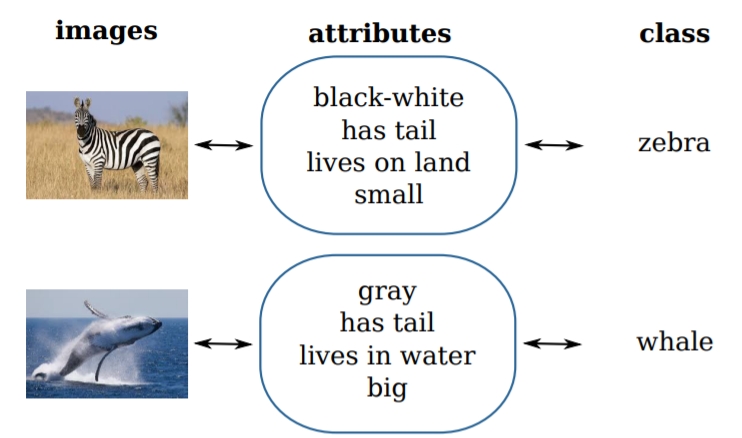}
\end{center}
\caption{Two examples of labeled images where the labels ``zebra" and ``whale" are described by four visual attributes. Examples taken from the CVPR 2016 Tutorial on Zero-shot learning for Computer Vision.}
\label{fig:semantic}
\end{figure}

Most importantly, pairwise learning methods are capable of generalizing to zero-shot problems, such as a novel target molecule in the drug design example, a novel tag in the document annotation example, or a novel course in the student grading example.
In recent years, specific zero-shot learning methods based on deep learning have become extremely popular in image classification applications. The central idea in all those methods is to construct semantic feature representations for class labels, for which various techniques might work. One class of methods constructs binary vectors of visual attributes \citep{Lampert2009,Palatucci2009,Liu2011,Fu2013}. Figure~\ref{fig:semantic} shows an example of two such vectors for two images. Another class of methods rather considers continuous word vectors that describe linguistic context of images \citep{Mikolov2013,Frome2013,Socher2013}. For the same two images of Figure~\ref{fig:semantic}, one could for example look up ``zebra" and ``whale" on Wikipedia, and extract a Word2Vec representation from this source.

Many zero-shot learning methods for image classification adopt principles that originate in kernel methods. The model structure can often be formalized as follows: 
\begin{equation}
\label{eq:zero-shot-pairwise}
f(\vec{x},\vec{t}) = \vec{w}^T \big(\phi(\vec{x}) \otimes \psi(\vec{t})\big)    
\end{equation}
This model in fact coincides with the primal formulation of (\ref{eq:zero-shot-pairwise}) with a Kronecker-based choice for $\Psi(\vec{x},\vec{t})$. Different optimization problems with this model have been proposed \citep{Frome2013,Akata2015,Akata2016}, and related methods provide nonlinear extensions \citep{Socher2013,Xian2016b}. Most of these optimization problems do not minimize squared error loss, and according to the definitions of Section~2, they should rather be seen as structured output prediction methods. Indeed, a representation such as (\ref{eq:pairwise}) is in fact commonly used in structured output prediction. SOP methods additionally have inference procedures that allow for finding the best-scoring targets in an efficient manner.

Some of the zero-shot learning methods from computer vision also turn out to be useful for the related field of text classification. For documents, it is natural to model a latent representation for both the (document) instances and class labels in a joint space \citep{Nam2016}.

\subsection{Representation-constructing methods}
\label{sec:methods-that-learn-target-representations}

In this section, we discuss methods that construct vector representations for targets without assuming side information as prior knowledge. So, in contrast to the methods that we discussed in Section \ref{sec:methods-that-use-target-representations}, no target representations need to be available. Instead, such representations are learned from scratch in the training phase\footnote{Let us remark that our notion of ``representation-constructing" differs substantially from the notion of ``representation learning" as commonly used in the area of deep neural networks. Here, we consider the construction of vector representations for targets. This is something that is not commonly done in multi-target prediction extensions of deep architectures.}. As a result, these methods are mainly useful for Settings B and C, while not being applicable to Setting D. In principle, target representations can also be learned for Setting A. However, since this involves additional complications, methods for Setting A will be discussed in Section~3.7. 

Methods that learn vector representations from scratch are in fact embedding-based methods. They all proceed from the assumption that the targets can be embedded in a vector space with a dimension lower than $m$. Principal component analysis (PCA), canonical correlation analysis (CCA), and partial least-squares (PLS) are simple techniques of that kind. In the case of PCA, one first embeds the targets in a low-dimensional space that is spanned by the most important principal components. In a second stage, one constructs regression models with the original features and the retained principal components as response variables. Predictions for new  observations are then obtained by feeding them to the regression models and applying the inverse transformation to the original space. CCA and PLS both solve a different formulation, in which targets and features are embedded jointly in a lower-dimensional space~\citep{Shawe-Taylor2004,Yu_et_al_2006}.  

One can also obtain an embedding in another way, by introducing a rank constraint on the parameter matrix $A$ during optimization:
$$\min_A ||Y - XA ||^2_F + \lambda \, \mathrm{rank}(A) 
$$
For example, \citet{Ando2005} and \citet{Chen2009} have shown that such an approach yields performance improvements in multi-task learning. Similar formulations for multivariate regression can be found in the statistical literature, starting in the seventies with well-known methods such as reduced-rank regression (RRR)~\citep{Izenman_1975} and FICYREG~\citep{Merwe_Zidek_1980}. According to~\citet{Breiman_Friedman_1997}, these methods have the same generic form:
$$
 \vec{f}^*(\vec{x}) = (T^{-1}GT)A\vec{x}\,  ,
$$
where $T$ is the matrix of sample canonical coordinates, the solution of canonical correlation analysis (CCA), and the diagonal matrix $G$ contains the shrinkage factors for scaling the solutions of ordinary linear regression $A$. Based on this equation, \citet{Breiman_Friedman_1997} also establish a close connection with stacking, which was discussed in Section~\ref{sec:methods-that-do-not-use-domain-knowledge}. 

While embedding-based methods originated in multivariate regression, they have been  used extensively for multi-label classification (as well as structured output prediction) in recent years. Nowadays they are very popular in the area of extreme multi-label classification, when the number of labels $m$ is extremely large. Then, embedding-based methods are needed for computational reasons, namely to obtain reasonable training and prediction times.  They mainly differ in the choice of compression or reduction technique used. One can distinguish methods based on PCA \citep{Weston_et_al_2002}, CCA \citep{Rai2009}, compressed sensing \citep{Hsu_et_al_2009}, singular value decompostion \citep{Tai_Lin_2010,Tai2012}, output codes \citep{Zhang2011},  landmark labels \citep{Balasu2012,Bi2013}, Bloom filters \citep{Cisse2013}, auto-encoders \citep{Wicker2016}, etc. In spite of their success, there is also evidence that embedding-based methods may have important shortcomings. In many extreme multi-label classification datasets, the low-rank assumption is violated as a result of the presence of many so-called tail labels. These are labels that appear very rarely in the dataset, sometimes less than five times in total, and recovering them with low-rank matrix approximation methods is very difficult \citep{Bhatia2015,Yen2016}.

\subsection{Matrix completion and hybrid methods}
\label{sec:matrix-completion-and-hybrid-methods}

In this section, we review matrix completion methods. In Section~2, such methods were claimed to be useful for an MTP setting with partially-observed matrices $Y$---in Figure~\ref{fig:foursettings} referred to as Setting A. Both the targets and the instances are observed, but not for all instance-target combinations. In Setting A, side information about instances or targets is not required per se. We hence distinguish between methods that ignore side information and methods that also exploit such information, in addition to analyzing the matrix $Y$. 

Inspired by the Netflix challenge in 2006, the former type of methods has been mainly popular in the area of recommender systems. Those methods often impute missing values by computing a low-rank approximation of the sparsely-filled matrix $Y$, and many variants exist in the literature, including algorithms based on nuclear norm minimization \citep{Candes2008}, Gaussian processes \citep{Lawrence2009}, probabilistic methods \citep{Shan2010}, spectral regularization \citep{Mazumder2010}, non-negative matrix factorization \citep{Gaujoux2010}, and alternating least-squares minimization \citep{Jain2013}. In addition to recommender systems, matrix factorization methods are commonly applied to social network analysis \citep{Menon2010}, biological network inference \citep{Gonen2012,Liu2015}, and travel time estimation in car navigation systems~\citep{Dembczynski_et_al_2013}. Let us stress that those methods differ substantially from the methods that were discussed in Section~\ref{sec:methods-that-learn-target-representations}. Here, matrix factorization is used to impute missing values in $Y$, while in Section~\ref{sec:methods-that-learn-target-representations} it was needed to reduce the dimensionality of the target space. In other words, here we are discussing methods that have been proposed for Setting A, whereas the methods from Section~\ref{sec:methods-that-learn-target-representations} are applicable to Setting B.    

In addition to matrix factorization, a few other methods exist for Setting A.  Historically, memory-based collaborative filtering has been popular, and corresponding methods are very easy to implement. They make predictions for the unknown cells of the matrix by modelling a similarity measure between either rows or columns---see e.g.\ \citep{Takacs2008}. For example, when rows and columns correspond to users and items, respectively, then one can predict novel items for a particular user by searching for other users with similar interests. In the resulting nearest neighbor search, the cosine similarity often outperforms other similarity measures. 

Many variants of matrix factorization and other collaborative methods have been presented, in which side information of rows and columns is considered during learning, in addition to exploiting the structure of the matrix $Y$---see e.g.\ \citep{Basilico2004,Abernethy2008,Adams2010,Fang2011,Zhou2011a,Menon2011,Zhou2012a}. One simple but effective method is to extract latent feature representations for instances and targets in a first step, and combine those latent features with explicit features in a second step, using any of the more conventional approaches that were discussed in the previous paragraphs \citep{Volkovs2012}. 
In light of Examples~1--3, when feature representations are available for proteins, documents, and students, it would be pointless to ignore them. Hybrid methods seek to combine the best of both worlds, by simultaneously modeling side information and the structure of $Y$. In addition to Setting A, they can often be applied to Settings B and C, which coincide, respectively, with a novel user and a novel item in recommender systems. In that context, one often speaks about cold-start recommendations.  

\section{Conclusion and future perspectives}

In this paper, we provided a unifying view on MTP problems and methods. In Section 2, we presented a general MTP framework and explained how well-known subfields of machine learning can be seen as specific instantiations of this framework. In Section 3, we gave an overview of MTP methods, categorized according to the prediction setting of interest and the domain knowledge that is available. For researchers who are new to the field of multi-target prediction, this categorization might help identify which method to use for a given MTP problem. As an overall conclusion, it should be clear that no MTP method is applicable, let alone optimal, under all conditions. It is up to the data scientist to choose the right method for a given problem, and we hope that this article will be helpful in this regard. 

There are several topics related to multi-target prediction that are not discussed in this paper. One important issue is the loss one seeks to optimize. As  briefly discussed in Section~3, we would like to emphasize that the methods reviewed in this paper are mainly useful for optimizing loss functions that are decomposable over targets---our example of mean squared error is exactly of that kind. For more complex loss functions, which mainly seem to appear in the multi-label classification literature, other algorithms will be needed. We refer to our previous work for an overview on this topic, with a particular focus on the subset zero-one loss, the rank loss, and the F1-measure \citep{Demb2012a,Waegeman2014}. 

We believe that research on multi-target prediction will be further intensified in the coming years, because novel applications are constantly emerging. Besides, several research questions still remain unanswered. In general, one can observe a trend towards analyzing larger target spaces, which require sophisticated algorithms that scale linearly or sublinearly in the number of targets. One also observes a trend towards extending classical machine learning paradigms for an MTP context---including semi-supervised learning, multi-instance learning, time series classification, data stream mining, network analysis, etc. We are convinced that extensions of that kind will be a driving force of fundamental machine learning research in the coming years. 

Finally, there is also a need for further theoretical research in multi-target prediction. Many of the recent algorithms rely on a purely heuristic or intuitive motivation, which is then tested (and usually confirmed) in empirical studies on classical benchmark datasets. Most of the time, however, a deeper understanding of those algorithms and a theoretical explanation of their behavior are missing. 
Needless to say, an understanding of that kind is needed to mature the field of multi-target prediction, and to unify the different branches discussed in this paper.

\section*{Appendix A: James-Stein Estimation}
\label{sec:James-Stein-Estimation}


In the late sixties, James and Stein discovered that the best estimator of the mean of a multivariate Gaussian distribution is not necessarily the maximum likelihood estimator. More formally, assume that $\theta$ is the unknown mean of a multivariate Gaussian distribution with dimension $m>2$ and a diagonal covariance matrix. Consider a single observation $\vec{y}$ randomly drawn from that distribution: 
$$\vec{y} \sim \mathcal{N}(\theta, \sigma^2 I) \,.$$ 
Using only this observation, the maximum-likelihood estimator for $\theta$ would be $\hat{\theta}_{ML} = \vec{y}$. James and Stein discovered that the maximum likelihood estimator is suboptimal in terms of mean squared error
$$\mathbb{E} \big[||\theta - \hat{\theta}||^2 \big] \,,
$$
where the expectation is over the distribution of $\vec{y}$. (In general, the expectation is taken over all samples that contain a single observation $\vec{y}$. Later on we will shortly discuss a situation in which we draw more than one observation to compute the value of the estimator).
An estimator with lower squared error can be obtained by applying a regularizer to the maximum likelihood estimator. In case $\sigma^2$ is known, the James-Stein estimator is defined as follows: 
$$\hat{\theta}_{JS} = \left(1 - \frac{(m-2)\sigma^2}{||\vec{y}||^2} \right) \vec{y} \,.$$
From a machine learning perspective, a regularizer is introduced that shrinks the estimate towards the zero vector, and hence reduces variance at the cost of introducing a bias. It has been shown that this biased estimator outperforms the maximum likelihood estimator in terms of mean squared error. 
The result even holds when the covariance matrix is non-diagonal, but in view of the discussion concerning target dependence, it is most remarkable for diagonal covariance matrices. In fact, in the latter case, it means that joint target regularization will be beneficial even if targets are intrinsically independent. This is somewhat in contradiction with what is commonly assumed in the machine learning literature. 

Let us notice, however, that the advantage of the James-Stein estimate over the maximum likelihood estimate will vanish for larger samples (of more than one observation). In the second term in parentheses, $\sigma^2$ is then divided by the size of the sample, so that the James-Stein estimate converges to the maximum likelihood estimate when the sample size grows to infinity.

The James-Stein paradox analyzes a very simple estimation setting, for which suboptimality of the maximum likelihood estimator can be proved analytically, but the principle extends to various multi-target prediction settings. By interpreting each component of $\theta$ as an individual target (and omitting the instance space, or reducing it to a single point), the maximum likelihood estimator coincides with independent model fitting, whereas the James-Stein estimator adopts a regularization mechanism that is very similar to most of the regularization techniques used in the machine learning literature. For some specific multivariate regression models, connections of that kind have been discussed in the statistical literature \citep{Breiman_Friedman_1997}. As long as mean squared error is considered as a loss function and errors follow a Gaussian distribution, one can immediately extend the James-Stein paradox to multivariate regression settings by assuming that target vectors $\vec{y}$ are generated according to the following statistical model:    
$$\vec{y} \sim \mathcal{N}(\theta(\vec{x}), \sigma^2 I) \, , 
$$ 
where the mean is now conditioned on the input space. For other loss functions, we are not aware of any formal analysis of that kind, but it might be expected that similar conclusions can be drawn.

\bibliographystyle{spbasic}   
\bibliography{referencesMTP}
\end{document}